%% file: main.tex
\documentclass[11pt,a4paper]{article}

\usepackage{acl}
\usepackage{times}
\usepackage{latexsym}
\usepackage[T1]{fontenc}
\usepackage[utf8]{inputenc}
\usepackage{amsmath}
\usepackage{amsfonts}
\usepackage{amssymb}
\usepackage{graphicx}
\usepackage{booktabs}
\usepackage{multirow}
\usepackage{colortbl}
\usepackage{microtype}
\usepackage{algorithm}
\usepackage{algorithmic}
\usepackage{xcolor}
\usepackage{pifont}
\usepackage{tabularx}
\usepackage[most]{tcolorbox}
\usepackage{subcaption}
\usepackage{adjustbox}

\title{Last Step Matters: Early Uncertainty Cannot Predict Failure in Long-Horizon Agents}

\author{
  \textbf{Zongyue Li}\textsuperscript{1,2}\thanks{Equal contribution.},
  \textbf{Chengyue Yu}\textsuperscript{2}\footnotemark[1],
  \textbf{Lei Zang}\textsuperscript{2}\footnotemark[1], \\
  \textbf{Chenyi Zhuang}\textsuperscript{2},
  \textbf{Linjian Mo}\textsuperscript{2},
  \textbf{Leilei Gan}\textsuperscript{1}\thanks{Corresponding author.} \\
  \textsuperscript{1}Zhejiang University
  \quad
  \textsuperscript{2}Ant Group \\
  \parbox{0.96\textwidth}{%
    \centering
    \texttt{zongyueli@zju.edu.cn, yuchengyue.ycy@antgroup.com} \\
    \texttt{zanglei.zl@antgroup.com, chenyi.zcy@antgroup.com} \\
    \texttt{linyi01@antgroup.com, leileigan@zju.edu.cn}
  }
}

\begin{document}
\maketitle

\begin{abstract}
Early failure prediction is important for long-horizon agents, as it enables timely intervention and can reduce inference and tool-use costs. Uncertainty quantification, such as verbal confidence and perplexity, offers a promising approach to detecting agent failures; however, it has not been explored whether these signals retain their discriminative power during the intermediate stages of long-horizon execution. We evaluate mainstream uncertainty signals on deep-research tasks and find that verbal confidence reliably distinguishes failures at trajectory completion, achieving a mean AUROC of 0.85, whereas all evaluated signals offer limited predictive value earlier in execution, with none exceeding a mean AUROC of 0.60 at 50\% trajectory progress. We identify an underlying mechanism explaining this gap: \emph{path switching}, where agents frequently abandon their current search direction in-trajectory, breaking the link between early signal and final outcome. These findings challenge the assumption that intermediate uncertainty can
reliably guide early intervention. They also motivate a practical recommendation for agent harnesses in deep-research settings: use final-step confidence to decide whether to restart, an approach that our experiments find more effective than in-trajectory intervention.
\end{abstract}

\begin{figure}[ht]
  \centering
  \includegraphics[width=0.95\columnwidth]{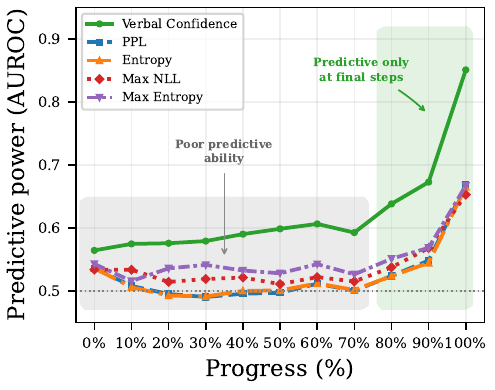}
  \caption{Predictive ability of uncertainty metrics for agent failure (measured by AUROC) across normalized trajectory progress. All metrics remain weak until 80--90\% progress, then rise sharply at completion (trajectories with $\geq$11 steps; AUROC averaged across five models and three deep research benchmarks). This pattern is consistent across the 15 combinations of models and benchmarks (Appendix~\ref{sec:appendix_norm_auroc_per_dataset}).}
  \label{fig:norm_auroc_intro}
\end{figure}

\begin{figure*}[ht]
  \centering
  \includegraphics[width=0.95\textwidth]{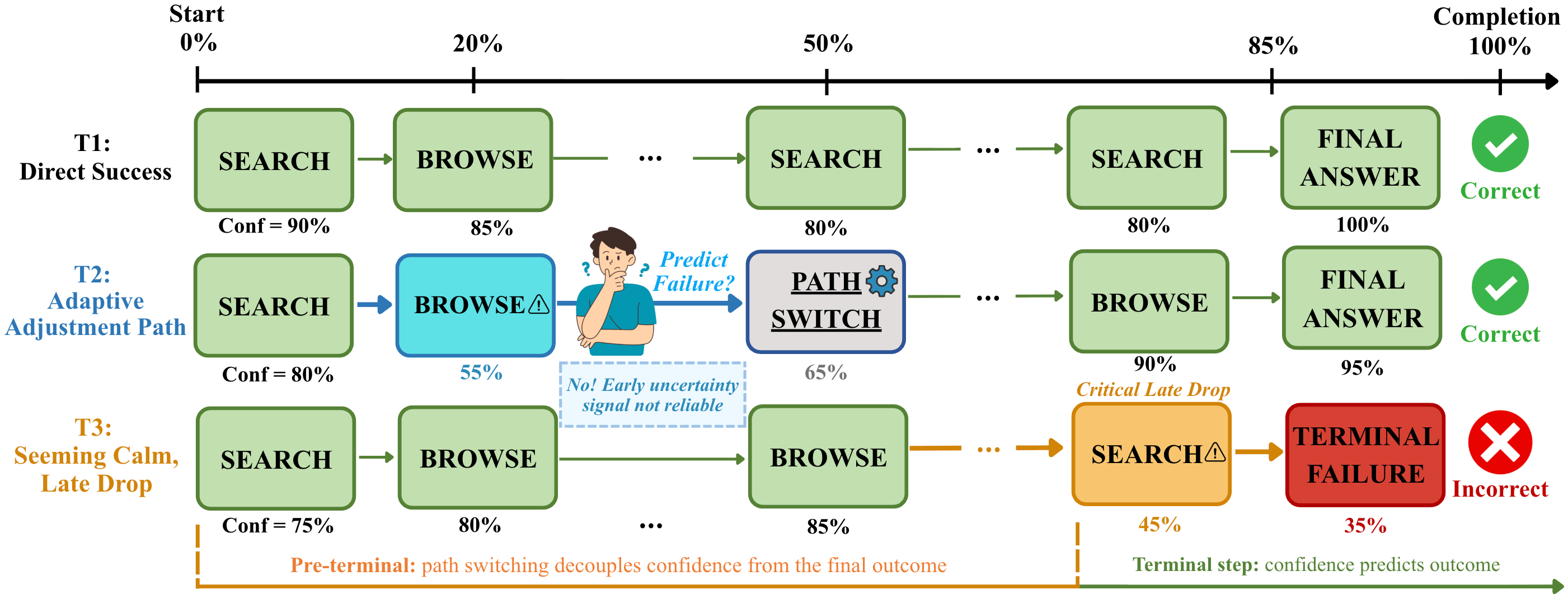}
  \caption{Illustration of three agent trajectories (T1, T2, and T3) showing how path switching can weaken the relationship between intermediate confidence and the final outcome. A low confidence signal at an intermediate stage (T2) does not indicate eventual failure when the agent subsequently switches to a new search direction, while a high confidence signal at an intermediate stage (T3) does not guarantee eventual success.}
  \label{fig:agent_failure_prediction}
\end{figure*}

\section{Introduction}
Large language model (LLM) agents can now autonomously tackle complex, long-horizon tasks through iterative information seeking, reasoning, and answer synthesis~\citep{yao2022react,schick2023toolformer,deng2023mind2web}. However, these agents remain prone to execution failures: they may get stuck in action loops~\citep{lu2025agentexit}, pursue incorrect reasoning paths~\citep{zhu2025wherefail}, or hallucinate information~\citep{huang2025hallucination}. Such failures are particularly costly in long-horizon tasks, as each additional
step incurs inference and tool-use costs~\citep{xiao2025improving,wang2025efficient}. Early failure prediction could therefore enable timely intervention before
these costs accumulate~\citep{lu2025agentexit}. Uncertainty quantification (UQ), including verbal confidence methods~\citep{tian-etal-2023-just,kadavath2022}, token probability methods~\citep{jelinek1977perplexity,shannon1948mathematical}, and consistency-based methods~\citep{farquhar2024detecting,manakul2023}, has emerged as a promising approach for predicting such failures~\citep{xia2025survey,ohuncertainty}. Prior work has shown that these metrics are discriminative in single-turn settings, while recent studies have extended their evaluation to multi-step agentic tasks~\citep{zhao-etal-2025-uncertainty,duan2025uprop,zhang2026htc}. However, existing studies generally focus on relatively short trajectories and
assess failure detection only at task completion. It therefore remains unclear
whether these metrics can reliably predict failures during \emph{long-horizon}
execution and how early they become informative. 

To fill this gap, we systematically evaluate uncertainty signals for failure
prediction throughout long-horizon agent execution. We consider three signal
families (verbal confidence, token-probability metrics, and consistency-based
metrics) across three deep-research benchmarks: BrowseComp~\citep{wei2025browsecomp},
BrowseComp-zh~\citep{zhou2025browsecompzh}, and Humanity's Last
Exam~\citep{phan2025hle}. Our experiments cover five models with parameter
counts ranging from 27B to 1T and address two questions:
\textbf{(1)~Effectiveness.} \emph{How reliably} can established uncertainty
metrics distinguish failed from successful trajectories at task completion?
\textbf{(2)~Timeliness.} More critically, \emph{when} do these signals become
informative? Can they identify failures early enough to support intervention
before task completion?

Our answer to the first question is encouraging: final-step verbal confidence
is a strong baseline, achieving an average AUROC (area under the receiver
operating characteristic curve) of 0.85 across the 15 model-benchmark combinations. The answer to the second question is sobering: all evaluated signals have \textbf{poor early predictive ability}, with none exceeding a mean AUROC of 0.60 at 50\% progress. As Figure~\ref{fig:norm_auroc_intro} shows, all metrics remain weak until 80--90\% progress, then rise sharply only upon completion.

We identify \textbf{path switching} as an underlying mechanism explaining this gap. Agents frequently abandon their current search direction during a trajectory. Figure~\ref{fig:agent_failure_prediction} illustrates how path switching weakens the link between intermediate confidence and the eventual outcome. After the final switch, verbal confidence becomes substantially more predictive. We also observe a pronounced separation near trajectory completion. Relative to the mean confidence over the first 70\% of each trajectory, final-step confidence is 15 percentage points higher for correct trajectories and 6 percentage points lower for incorrect trajectories. These findings indicate that intermediate uncertainty has limited utility for early failure prediction. Our intervention experiments further show that confidence-guided post-completion retries outperform in-trajectory intervention.

Our contributions are threefold: \textbf{(1)}~We provide the first
\textbf{systematic study} of how the predictive ability of uncertainty signals
changes throughout long-horizon agent trajectories in deep-research tasks. Our
evaluation spans three representative UQ families and distinguishes failure
\emph{detection} at completion from failure \emph{prediction} before completion.
\textbf{(2)}~We identify \textbf{path switching} as a behavioral phenomenon that
helps explain the weak predictive ability of early uncertainty signals.
\textbf{(3)}~We derive \textbf{practical recommendations for agent harnesses}:
our results favor post-completion retries guided by final-step confidence over
in-trajectory interventions triggered by intermediate confidence.

\section{Theoretical Background}

Following~\citet{ohuncertainty}, we frame uncertainty in LLM agents at two levels. At the step level, the uncertainty of the agent's action distribution is defined as:
\begin{equation}
\label{eq:slu}
\text{SLU}_t(x) = U\!\left[P(a_t \mid e_{t-1}, o_{t-1})\right]
\end{equation}
where $x$ denotes the task input, $a_t$ denotes the action at step $t$, and $e_{t-1}$ and $o_{t-1}$ denote the environment state and observation, respectively. $U(\cdot)$ is a generic uncertainty measure (e.g., entropy $H[P]$, perplexity $\exp H[P]$, or $100{-}\text{conf}_t$ for verbal confidence). At the trajectory level, step-level values are aggregated into a scalar:
\begin{equation}
\label{eq:tlu}
\text{TLU}(\tau) = \mathcal{A}\!\left[\text{SLU}_1(x),\, \ldots,\, \text{SLU}_T(x)\right]
\end{equation}
where $\tau$ denotes the trajectory generated for input $x$, and $\mathcal{A}$ may be the mean, last-step value, tail-weighted mean, or other aggregation operators.

\section{Experimental Setup}

\subsection{Datasets and Models}

We evaluate on three benchmarks (Table~\ref{tab:datasets}). \textbf{BrowseComp}~\citep{wei2025browsecomp} is a web-browsing question-answering benchmark comprising 1{,}266 questions that require persistently navigating the internet to find hard-to-find, entangled information. \textbf{BrowseComp-zh}~\citep{zhou2025browsecompzh} is its Chinese-language counterpart. \textbf{HLE}~\citep{phan2025hle} is a multimodal benchmark of 2{,}500 expert-level questions spanning mathematics, humanities, and natural sciences. Following \citet{parallelmuse2025}, we use the full BrowseComp-zh set (289 questions) and a randomly sampled subset of BrowseComp (200 questions); for HLE, we use only the text-only subset (157 questions).

We evaluate open-weight models spanning 27B to 1T parameters (Table~\ref{tab:models})~\citep{qwen3.5,5team2025glm45agenticreasoningcoding,zai2025glm47,deepseekv32,kimi26blog}. We restrict our evaluation to open-weight models because computing entropy and related uncertainty metrics requires token-level probability distributions, which are unavailable for closed-weight models.

Correctness is determined by comparing the agent's final answer to the ground truth using an LLM as a judge~\citep{gu2024survey}.

\begin{table}[ht]
  \centering
  \footnotesize
  \resizebox{\columnwidth}{!}{%
  \begin{tabular}{lrrrr}
  \toprule
  \textbf{Dataset} & \textbf{\# Tasks} & \textbf{Acc.} & \textbf{Avg Steps} & \textbf{Max Steps} \\
  \midrule
  BrowseComp & 200 & 32.0\% & 31.6 & 80 \\
  BrowseComp-zh & 289 & 49.8\% & 20.7 & 80 \\
  HLE & 157 & 34.1\% & 13.5 & 80 \\
  \bottomrule
  \end{tabular}%
  }
  \caption{Dataset statistics.}
  \label{tab:datasets}
  \small
  \begin{tabular}{lccc}
  \toprule
  \textbf{Model} & \textbf{Total} & \textbf{Activated} & \textbf{Arch.} \\
  \midrule
  Qwen3.5-27B & 27B & 27B & Dense \\
  Qwen3.5-122B-A10B & 122B & 10B & MoE \\
  GLM-4.7 & 355B & 32B & MoE \\
  DeepSeek-V3.2 & 671B & 37B & MoE \\
  Kimi K2.6 & 1T & 32B & MoE \\
  \bottomrule
  \end{tabular}
  \caption{Evaluated models, spanning 27B to 1T total parameters. All are open-weight models, allowing access to token probabilities throughout each trajectory.}
  \label{tab:models}
  \end{table}

\input{tab_cross_family}

\subsection{Agentic Setting}

We follow the agent setup of \citet{browseconf2026}. The agent has access to two tools: (1)~\textbf{Search}, which queries the Google search engine and returns the top 10 results per query (each with a title, snippet, and URL), and (2)~\textbf{Visit}, which extracts the full content of designated webpages via Jina\footnote{\url{https://jina.ai/}} and then summarizes goal-relevant information. No memory compression, context folding, or other auxiliary tools are used. All models are configured with a temperature of 1.0, a top-$p$ value of 0.95, a context length of 128K tokens, and a maximum of 80 tool calls per trajectory. We perform eight rollouts per task for each model across all three benchmarks.

\subsection{Uncertainty Metrics}
\label{sec:metrics}

We evaluate the following families of uncertainty metrics:

\paragraph{Verbal confidence.} Following the practice of prompting models to verbally state their uncertainty~\citep{tian-etal-2023-just,wei2024simpleqa}, we extend this to the agentic setting by requiring the agent to output a confidence score (0--100) at each step, producing a verbal confidence sequence over the trajectory.

\paragraph{Token probability metrics.} We compute four metrics from log probabilities at each step: perplexity (PPL)~\citep{jelinek1977perplexity}, maximum token negative log-likelihood (Max NLL)~\citep{manakul2023}, Shannon entropy~\citep{shannon1948mathematical}, and maximum token entropy~\citep{manakul2023}.

\paragraph{Metric aggregation.}
All metrics above are computed at each step, producing a sequence for each
trajectory. We reduce each sequence to a scalar using six operators:
\texttt{mean}, \texttt{running max}, \texttt{running min}, \texttt{last step},
\texttt{last-3 mean}, and \texttt{last-5 mean}. Applying these six operators to
the five base metrics yields 30 scalar features.

\paragraph{Consistency-based metrics.} Consistency-based metrics~\citep{xia2025survey} require multiple rollouts per task and quantify uncertainty from the variation across answers or trajectories. These metrics are not suitable for our goal of early prediction: they require $N{\times}$ compute for $N$ rollouts and rely on completed answers or trajectories, preventing them from being computed from partial trajectories. We nevertheless include ten such metrics (diversity, semantic entropy, predictive entropy, etc.) in Table~\ref{tab:consistency} for completeness, with abbreviations and brief descriptions in Appendix~\ref{sec:appendix_consistency_metrics}.

\subsection{Evaluation Protocol}
\label{sec:eval_protocol}

\paragraph{AUROC.}
We use the area under the ROC curve (AUROC) to measure the predictive ability
of each uncertainty metric. An AUROC of 0.5 indicates chance-level
discrimination, whereas an AUROC of 1.0 indicates perfect discrimination. At
each trajectory progress point, we compute AUROC separately for each of the 15
model--benchmark combinations and report their mean.

\paragraph{Backward truncation analysis.}
Since trajectories vary in length, we first normalize trajectory progress into
11 equally spaced points (0\%, 10\%, 20\%, \ldots, 100\%). We restrict the
analysis to trajectories with at least 11 steps to ensure adequate coverage of
these progress points. At each point, we use only the information available up
to that point to predict the final trajectory outcome. We then compute the
AUROC of each uncertainty signal across trajectories, revealing how its
predictive ability changes over the course of execution.

\section{Discriminative Ability of Uncertainty Metrics}
\label{sec:discriminative_power}

\subsection{Full-Trajectory Failure Detection}
\label{sec:cross_family}

Our first question is whether uncertainty quantification remains effective for
failure detection in long-horizon agents when the full trajectory is available.
We find that it does: final-step verbal confidence achieves a mean AUROC of
0.85 across the 15 model--benchmark combinations. Table~\ref{tab:cross_family}
summarizes the results, with complete results provided in
Appendix~\ref{sec:cross_family_full}.

These results suggest that the model itself serves as both an effective
uncertainty estimator and a temporal information aggregator: its verbal
confidence is more discriminative than token-probability metrics, while its
final-step assessment outperforms hand-designed temporal aggregators such as
\texttt{mean} and \texttt{max}.

Among consistency metrics (Table~\ref{tab:consistency}), answer-frequency- and
graph-based metrics (diversity and degree uncertainty) generally outperform
log-probability-based metrics, but still trail verbal confidence
overall. The sole exception is HLE, where discrete semantic entropy outperforms
verbal confidence for both evaluated models.

\subsection{Early Failure Prediction}
\label{sec:early_detection}

We next ask whether failures can be identified before the agent reaches its
conclusion. Figure~\ref{fig:norm_auroc_intro} shows how the failure-prediction
ability of five uncertainty metrics, measured by AUROC, changes over trajectory
progress. We evaluate intermediate stages using the backward truncation
protocol (\S\ref{sec:eval_protocol}). At each progress point, we report the mean
AUROC across the 15 model--benchmark combinations. 

Predictive ability remains
limited: at 50\% progress, no metric exceeds a mean AUROC of 0.60; even at
90\%, all metrics remain below 0.70. Verbal confidence performs best throughout,
but its mean AUROC rises sharply to 0.85 only at completion.

\begin{figure}[ht]
  \centering
  \includegraphics[width=0.95\columnwidth]{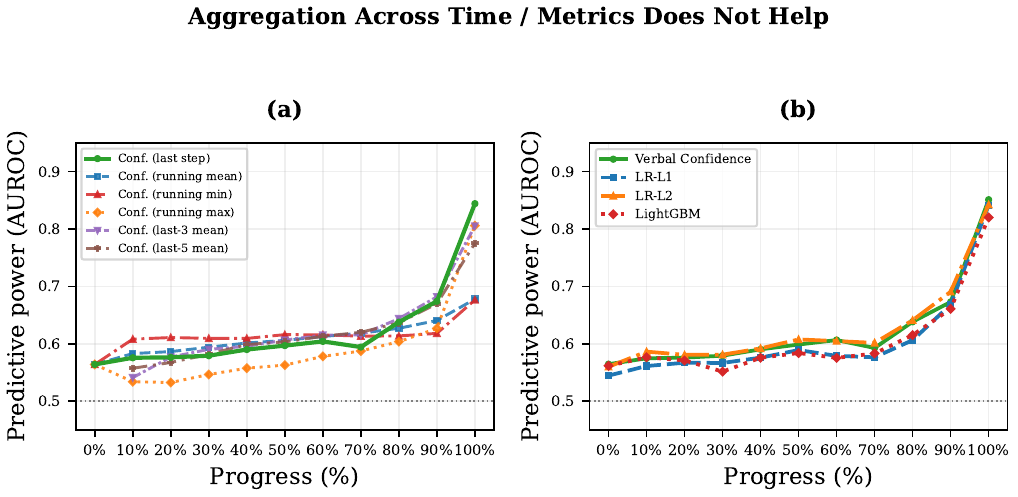}
  \caption{Temporal aggregation and feature combination do not improve early prediction. (a)~Verbal confidence with six temporal aggregation methods; running minimum provides only a small improvement in early AUROC. (b)~Classifiers combining five uncertainty metrics do not outperform verbal confidence at any progress point.}
  \label{fig:norm_auroc_classifier}
\end{figure}

\subsection{Can Feature Aggregation/Combination Help?}
\label{sec:nonlinear}

We ask whether aggregating verbal confidence over time or combining multiple
uncertainty metrics can improve early prediction.
Figure~\ref{fig:norm_auroc_classifier}(a) compares six temporal aggregation
methods for verbal confidence. At 50\% trajectory progress, running minimum
performs best, but improves mean AUROC only modestly (0.60 to 0.62). At completion, no temporal
aggregation method outperforms final-step confidence.
Figure~\ref{fig:norm_auroc_classifier}(b) tests whether combining five
uncertainty features improves early prediction using L1- and L2-regularized
logistic regression and LightGBM~\citep{ke2017lightgbm}, evaluated with
five-fold stratified cross-validation. None of the combined models outperforms
verbal confidence alone at any progress point, showing that feature combination
provides no benefit for early prediction.

\section{Path Switching and Poor Early Predictive Ability}
\label{sec:micro_mechanisms}

\subsection{Path Switching}
\label{sec:implicit_restart}

We observe frequent \emph{path switching}, in which an agent abandons its
current search direction during a trajectory. This behavior may help explain
the limited predictive ability of intermediate uncertainty signals.

Table~\ref{tab:case_study} illustrates this with an example trajectory from BrowseComp. The agent first searches by the ``ambassador'' constraint alone (Steps 1--2), then switches to the ``Nobel Prize'' constraint (Step~3), and then further refines by combining ``ambassador'' and ``Nobel Prize'' (Step~4). It is only after this second switch, which synthesizes all three key constraints (headmaster, Nobel laureate, ambassador), that the search yields a productive lead. The pre-switch confidence scores
(70\% and 65\%) reflect the agent's assessment of partial search directions
that are later abandoned and thus provide little indication of eventual
success.

\input{tab_consistency}

\begin{table}[ht]
\centering
\small
\resizebox{\columnwidth}{!}{%
\begin{tabular}{clrcp{4.2cm}}
\toprule
\multicolumn{5}{p{7.2cm}}{\textbf{Q:} Who was headmaster of a school in the 1940s--60s, taught a future Nobel laureate, and later became ambassador in the 1970s?} \\
\multicolumn{5}{p{7.2cm}}{\textbf{A:} Francis L.\ Bartels} \\
\midrule
\textbf{Step} & \textbf{Tool} & \textbf{Conf} & \textbf{Sw.} & \textbf{Search query / Action} \\
\midrule
1 & search & 70\% & & headmaster turned ambassador 1970s \\
2 & search & 65\% & & former headmaster became ambassador 1970s \\
3 & search & 60\% & $\leftarrow$ & headmaster 1940s 1960s student won \textbf{Nobel Prize} \\
4 & search & 55\% & $\leftarrow$ & Nobel laureate attended school headmaster \textbf{ambassador} \\
5 & visit & 85\% & & Francis L.\ Bartels (Wikipedia) \\
6 & search & 90\% & & Mfantsipim School alumni Nobel Prize winner \\
7 & visit & 95\% & & Confirm answer \\
\bottomrule
\end{tabular}%
}
\caption{Example of path switching in a BrowseComp trajectory. Sw.\ = path switch. Confidence before the switch reflects search directions that are later abandoned.}
\label{tab:case_study}

\vspace{4pt}
\centering
\small
\resizebox{\columnwidth}{!}{%
\begin{tabular}{lrr}
\toprule
\textbf{Benchmark} & \textbf{\% traj.\ with switch} & \textbf{Avg \# switches / traj.} \\
\cmidrule(lr){2-3}
\multicolumn{3}{l}{\itshape Detected by keywords such as ``try a different approach''} \\
\midrule
BrowseComp & 86.8\% & 8.5 \\
BrowseComp-zh & 69.7\% & 4.7 \\
HLE & 36.9\% & 1.3 \\
\bottomrule
\end{tabular}%
}
\captionof{table}{Prevalence of path switches.}
\label{tab:switch_stats}
\end{table}

\paragraph{Path switch detection.}
We detect switches using 24 regular expressions matching explicit changes of
direction, such as ``let me try a different approach.'' To validate the
detector, we manually annotate 200 BrowseComp trajectories generated by
GLM-4.7 and compare the detected switch steps against human annotations. The
detector achieves 91.58\% accuracy, 99.93\% precision, 75.91\% recall, and
86.28\% F1, making it a high-precision but conservative proxy for path
switching.

\paragraph{Prevalence of path switching.}
Path switching is prevalent across all three benchmarks: 86.8\%, 69.7\%, and
36.9\% of trajectories contain a detected switch on BrowseComp, BC-zh, and
HLE, respectively, with averages of 8.5, 4.7, and 1.3 switches per trajectory
(Table~\ref{tab:switch_stats}).

\begin{figure}[ht]
\centering
\begin{minipage}[b]{0.48\columnwidth}
  \centering
  \includegraphics[width=\linewidth]{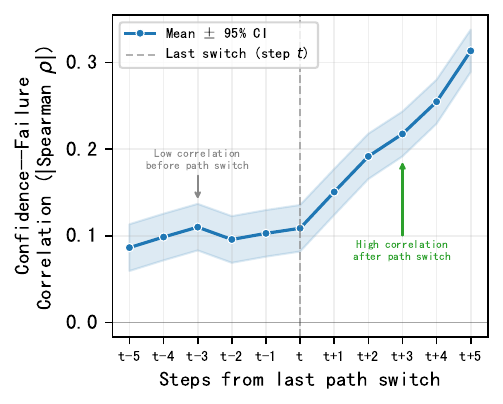}
\end{minipage}
\hfill
\begin{minipage}[b]{0.48\columnwidth}
  \centering
  \includegraphics[width=\linewidth]{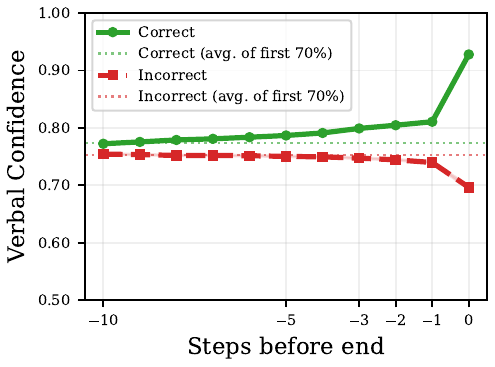}
\end{minipage}
\caption{Path switching and poor early predictive ability. (a)~Absolute Spearman correlation between confidence and correctness before and after the last detected path switch (step~$t$). (b)~Confidence surge relative to the average over the first 70\% of the trajectory. Correct trajectories exhibit a confidence increase, while incorrect trajectories decline slightly.}
\label{fig:path_switch_combined}
\end{figure}

\paragraph{Predictive ability around path switches.}
If path switches weaken the relationship between intermediate uncertainty and
the final outcome, uncertainty metrics should be less predictive before a
switch than afterward. Figure~\ref{fig:path_switch_combined}(a) tests this
prediction using $|\text{Spearman } \rho|$ between confidence and correctness
at each position relative to the last detected switch (step $t$). Before the
switch (steps $t{-}5$ to $t{-}1$), $|\rho|$ remains below 0.15 across all
model--benchmark combinations; afterward, it rises monotonically to 0.3--0.5
by step $t{+}5$. To account for task difficulty as a potential confounder, we
repeat the analysis within difficulty strata and observe the same pattern in
each group (Appendix~\ref{sec:appendix_switch_difficulty}).

\subsection{Post-Switch Confidence Surge}
\label{sec:terminal_kick_subsec}

Frequent path switching may delay reliable confidence assessment
until the agent settles on its final direction. We therefore expect confidence
to become informative only near trajectory completion, rising on successful
trajectories but not on unsuccessful ones.

We quantify this \textbf{confidence surge} as the difference between final-step
verbal confidence and mean verbal confidence over the first 70\% of the
trajectory. Figure~\ref{fig:path_switch_combined}(b) supports this prediction:
confidence increases by 15 points on average for correct trajectories but
decreases by 6 points for incorrect trajectories. The dashed lines
indicate each group's mean confidence over the first 70\% of the trajectory.
This late-stage divergence appears across all 15 model--benchmark combinations
(Appendix~\ref{sec:appendix_confidence_trajectories}).

Together, these findings support a path-switching account of the
temporal gap in predictive ability: confidence reflects transient search paths
early in execution and becomes aligned with final correctness only after the
agent settles on a direction near trajectory completion.

\section{Discussion}

\subsection{Sensitivity to Confidence Elicitation}

In our main experiments, confidence is collected alongside each tool call (Appendix~\ref{sec:appendix_prompts}). A natural concern is whether the observed weak early-stage predictive power is an artifact of this specific elicitation method. We evaluate two alternatives: \textbf{In-content}, where the model reports its confidence in \texttt{<confidence>} tags after its reasoning and before the tool call; and \textbf{Two-stage}, where confidence is obtained through a separate model call after each step~\citep{tian-etal-2023-just}.

Results for Qwen3.5-122B-A10B, averaged across two benchmarks (BrowseComp and BrowseComp-zh) are shown in Figure~\ref{fig:sensitivity_combined}(a). All three elicitation methods exhibit the same qualitative trend: weak predictive performance early in the trajectory, followed by a sharp increase toward completion.

\subsection{Calibration of Verbal Confidence}
\label{sec:calibration}

LLMs tend to be overconfident when asked to report their confidence as a
numerical value~\citep{wei2024simpleqa}. A natural concern is whether such
miscalibration affects our main findings.

We indeed find substantial overconfidence in long-horizon tasks at both intermediate and final stages. At 50\% trajectory progress, mean confidence is 77.05\%, compared with an actual success rate of 30.77\%, and the expected calibration error (ECE) is 0.4629~\citep{guo2017calibration}. At trajectory completion, mean confidence decreases to 64.99\%, while the ECE falls to 0.3422. Thus, although final-step confidence is better calibrated than intermediate confidence, substantial overconfidence remains.

However, this miscalibration does not affect our main conclusions regarding discriminative power. We measure discriminative power using AUROC, which depends on the relative ranking of confidence scores rather than their absolute values. Therefore, any monotonic recalibration that preserves this ranking leaves the AUROC unchanged.

On the other hand, recalibration may affect precision and recall computed at fixed absolute confidence thresholds. Figure~\ref{fig:precision_recall} reports these metrics at thresholds of 0.8 and 0.9. To ensure that our conclusions do not depend on these particular choices, Appendix~\ref{sec:appendix_pr_curves} presents full precision--recall curves obtained by sweeping the threshold over its entire range. Failure-detection performance at 50\% trajectory progress is consistently and substantially worse than at 100\% regardless of the threshold chosen.

\subsection{Beyond Deep Research Tasks}
\label{sec:beyond_deep_research}

The weak association between intermediate uncertainty signals and final outcomes may partly depend on whether agent actions produce persistent changes to the environment. In our deep-research tasks, search and page-visit actions
are largely read-only, allowing an agent to abandon one search direction and explore another without lasting consequences. Coding tasks, by contrast, are
more path-dependent: commands can create files, modify code, or otherwise alter the system state, causing earlier actions to constrain subsequent ones. Such persistence increases the cost of changing direction and may make intermediate
uncertainty signals more predictive of the eventual task outcome.

\begin{figure}[ht]
\centering
\includegraphics[width=\columnwidth]{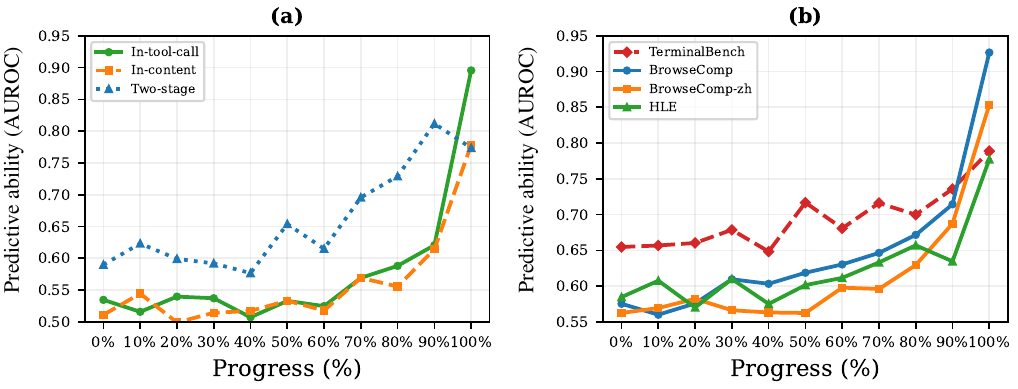}
\caption{Sensitivity analysis. (a)~Sensitivity to confidence elicitation: predictive ability (AUROC) by normalized progress, comparing three elicitation methods. All methods show the same qualitative trend of poor predictive ability early followed by a confidence surge. (b)~Sensitivity to task type: coding tasks (TerminalBench~2.0) vs.\ deep research benchmarks. Coding tasks show higher intermediate AUROC.}
\label{fig:sensitivity_combined}
\end{figure}

To examine this hypothesis, we first compare the predictive ability of intermediate confidence across the two task types while holding the underlying model fixed. Figure~\ref{fig:sensitivity_combined}(b) contrasts the performance of GLM-4.7 on TerminalBench 2.0~\citep{merrill2026terminalbench} with its performance on the three deep-research benchmarks. TerminalBench 2.0 comprises 89 tasks in realistic command-line environments. In each trajectory, the agent executes shell commands, observes their outputs, and reports a confidence score at every step. Across intermediate progress points, GLM-4.7 achieves AUROC values of 0.66--0.72 on TerminalBench, compared with 0.56--0.62 on the deep-research tasks. At trajectory completion, its AUROC reaches 0.79 on TerminalBench, compared with 0.78--0.93 across the deep-research benchmarks. These results are consistent with the hypothesis that intermediate confidence retains more predictive information in state-changing, path-dependent coding environments.

To determine whether this pattern extends beyond GLM-4.7, we further evaluate two Qwen3.5 models on TerminalBench. As shown in Table~\ref{tab:terminalbench_models}, both models exhibit substantial intermediate predictive power, with AUROC values ranging from 0.655 to 0.781 across the evaluated progress points and exceeding 0.85 at completion. Together with the GLM-4.7 results, these findings show a consistent pattern across all three models: intermediate confidence is more predictive on coding agents benchmarks than on the deep-research benchmarks.

As a behavioral check on the proposed path-dependence explanation, we compare how frequently agents switch between solution paths across task types. Path switching is substantially less frequent on TerminalBench than on BrowseComp (Appendix~\ref{sec:appendix_switch_environments}), consistent with the idea that persistent state changes constrain changes in direction during coding tasks. The alignment between reduced path switching and higher intermediate AUROC provides supporting evidence that path dependence may make intermediate confidence more informative.

\begin{table}[t]
\centering
\small
\resizebox{\columnwidth}{!}{%
\begin{tabular}{lrrrr}
\toprule
\textbf{Model} & \textbf{20\%} & \textbf{50\%} & \textbf{70\%} & \textbf{100\%} \\
\midrule
GLM-4.7 & 0.660 & 0.717 & 0.716 & 0.789 \\
Qwen3.5-122B-A10B & 0.760 & 0.781 & 0.781 & 0.866 \\
Qwen3.5-27B & 0.674 & 0.655 & 0.710 & 0.851 \\
\bottomrule
\end{tabular}%
}
\caption{Verbal confidence AUROC across normalized trajectory progress on TerminalBench~2.0. At 50\% progress, AUROC ranges from 0.655 to 0.781 across the three models.}
\label{tab:terminalbench_models}
\end{table}

\section{Implications for Agent Harnesses}
\label{sec:implications}

What implications do our findings carry for the design of agent harnesses? Our findings suggest that intermediate uncertainty may offer limited benefit for guiding intervention in these tasks, for two reasons. First, intermediate uncertainty signals have poor predictive ability for the final outcome. Second, since the agent already revises its search direction through path switching, additional intervention based on intermediate uncertainty seems redundant.

\begin{figure}[ht]
  \centering
  \includegraphics[width=0.9\columnwidth]{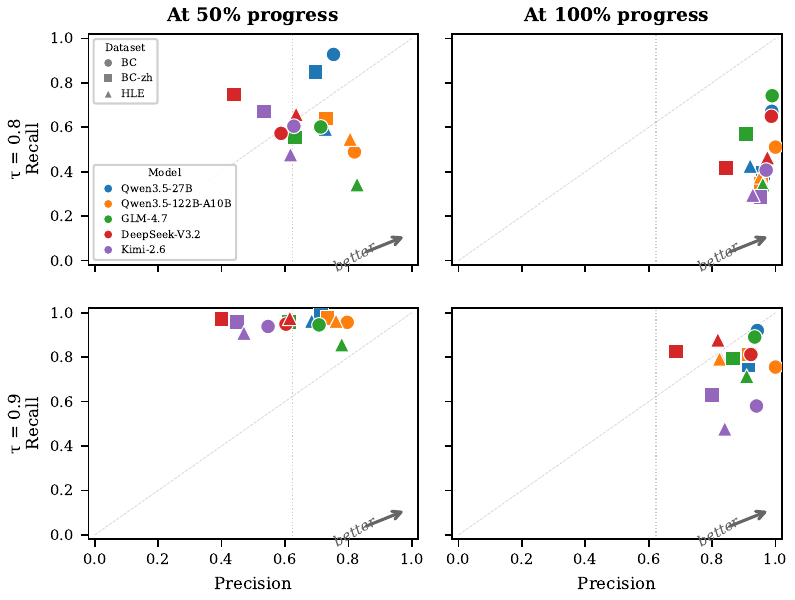}
  \caption{Failure detection precision and recall at two confidence thresholds ($\tau{=}0.8$ and $\tau{=}0.9$; predict failure when confidence $< \tau$). Columns compare 50\% vs.\ 100\% trajectory progress. At 50\%, performance is limited at both thresholds; at 100\%, the precision--recall tradeoff improves. Each point is one combination of model and benchmark. Full precision--recall curves in Appendix~\ref{sec:appendix_pr_curves}.}
  \label{fig:precision_recall}
\end{figure}

To make this concrete, Figure~\ref{fig:precision_recall} shows the
precision and recall of failure detection at two confidence
thresholds ($\tau{=}0.8$ and $\tau{=}0.9$), evaluated at 50\% and 100\%
trajectory progress and averaged across all 15 model--benchmark
combinations. At 50\% progress, performance is limited at both
thresholds: $\tau{=}0.8$ yields a precision of 0.68 and a recall of 0.62,
whereas $\tau{=}0.9$ achieves a recall of 0.96 but a precision of only 0.64,
only slightly above the average failure rate of 0.62. Neither
threshold provides acceptable performance in real-world settings. At 100\%
progress, the picture improves: $\tau{=}0.8$ achieves a precision
of 0.95 and a recall of 0.46, whereas $\tau{=}0.9$ achieves a precision of 0.88
and a recall of 0.76, indicating that confidence signals are useful
for guiding interventions primarily at task completion.

\begin{figure}[ht]
  \centering
  \includegraphics[width=\columnwidth]{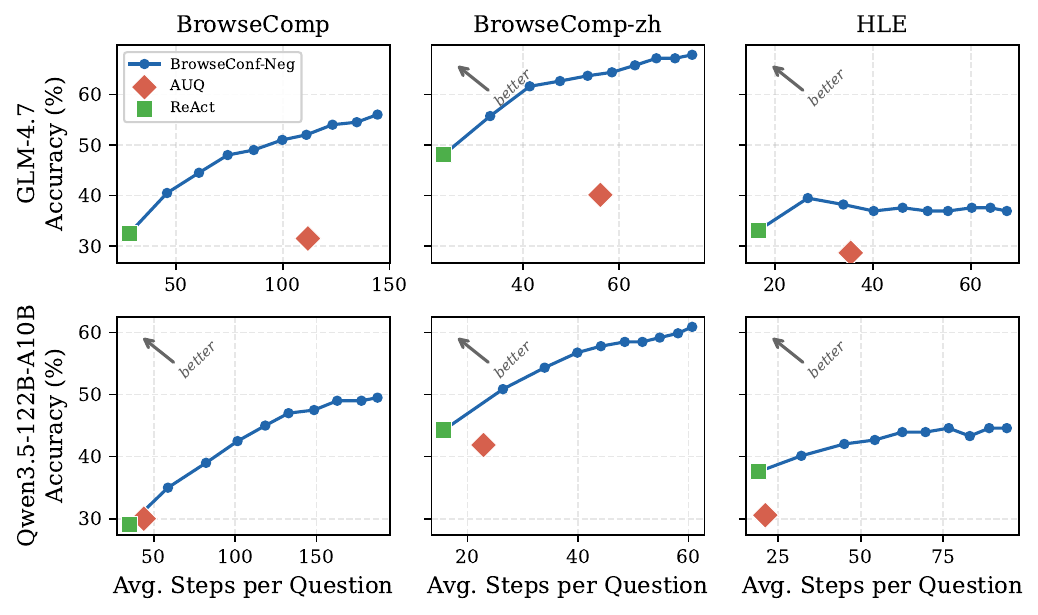}
  \caption{Agent harness: accuracy vs.\ efficiency. Post-completion intervention (BrowseConf-Neg) Pareto-dominates In-trajectory intervention (AUQ) across both evaluated models and all three benchmarks. Upper-left is better (fewer steps, higher accuracy).}
  \label{fig:pareto_auq_browseconf}
\end{figure}

Existing work has already explored confidence-based intervention at two distinct points in the agent loop:
\begin{itemize}
  \item \textbf{In-trajectory intervention} (AUQ; \citealt{zhang2026auq}): monitors verbal confidence at each step; when it drops below a threshold, the agent pauses to reflect and replan within the current trajectory, consuming additional steps before continuing.
  \item \textbf{Post-completion intervention} (BrowseConf-Neg; \citealt{browseconf2026}): waits until the agent finishes a full trajectory, then checks the final verbal confidence; if it falls below a threshold, the agent restarts the entire task while informed that the previous answer was likely incorrect.
\end{itemize}

We reproduce both methods on our benchmarks. ReAct (no intervention) serves as the shared baseline. AUQ is configured with the reflection threshold $\tau{=}0.8$ (as in the original paper) and a best-of-$N$ reflection budget, where the agent keeps trying until confidence rises above the threshold or the maximum step count is reached. BrowseConf-Neg is run with up to 10 total rollouts, restarting from scratch whenever the final confidence falls below $\tau{=}0.9$ (as in the original paper), and conditioning each retry on the negation of the prior low-confidence answer.

Figure~\ref{fig:pareto_auq_browseconf} compares the two strategies in the accuracy-efficiency plane. Each BrowseConf-Neg point corresponds to a rollout budget ranging from 1 to 10. The horizontal axis measures \emph{effective steps}, including regular agent steps, AUQ's in-trajectory reflection steps, and all steps from BrowseConf-Neg retry. Thus, points with similar horizontal coordinates represent comparable average interaction budgets.

Post-completion intervention (BrowseConf-Neg) Pareto-dominates in-trajectory intervention (AUQ) across both evaluated models and all three benchmarks: with comparable step budget, it achieves higher accuracy. Furthermore, its accuracy rises steadily with each additional retry, showing that this specific confidence-guided restart policy converts additional compute into accuracy gains.

\subsection{Recommendations for Agent Harnesses}

Based on the analysis above, we offer three recommendations for agent harnesses in deep-research settings:
\textbf{(1)~Use final-step verbal confidence to identify failures.} Verbal confidence reliably distinguishes failures at trajectory completion but offers limited predictive value earlier in execution. \textbf{(2)}~\textbf{When final-step confidence is low, restart with prior answer awareness}~\citep{browseconf2026}. Intervention based on intermediate uncertainty may incur false-positive costs and disrupt path switching that
would otherwise allow the agent to recover. \textbf{(3)~Move beyond standard uncertainty quantification for early failure detection.} Standard uncertainty signals have limited predictive power early in a trajectory. Reliable early failure detection may therefore require richer trajectory-aware methods or broader harness designs, such as self-critique or multi-agent monitoring. Our illustrative external-monitor experiment provides preliminary support for this broader direction (Appendix~\ref{sec:appendix_external_monitor}).

\section{Related Work}

Verbal confidence methods~\citep{tian-etal-2023-just}, methods based on token probabilities~\citep{kadavath2022}, and consistency-based methods~\citep{farquhar2024detecting,manakul2023} have been widely studied in single-turn settings. Recent work on uncertainty quantification for LLM agents~\citep{ohuncertainty} spans two related themes: quantifying uncertainty over agent trajectories and using uncertainty to guide in-trajectory interventions or post-completion retries.

For uncertainty quantification, SAUP~\citep{zhao-etal-2025-uncertainty} aggregates step-level uncertainty using situation-dependent weights, while UProp~\citep{duan2025uprop} distinguishes uncertainty inherent in the current decision from uncertainty propagated by preceding decisions. By incorporating intermediate uncertainty into a trajectory-level estimate, both methods implicitly assume that intermediate steps contribute useful information to the final prediction, but neither directly evaluates this assumption by measuring predictive ability throughout the trajectory. Agentic Confidence Calibration~\citep{zhang2026htc} finds that final-step token confidence can outperform full-trajectory averaging, but does not empirically evaluate early prediction during execution.

Prior work has also used uncertainty signals to guide agent execution. BrowseConf~\citep{browseconf2026} uses final-step verbal confidence to guide retry decisions. ParallelMuse~\citep{parallelmuse2025} partitions trajectories into functional regions and uses uncertainty to guide path reuse and branching during partial rollouts. AUQ~\citep{zhang2026auq} monitors intermediate verbal confidence and triggers reflection during the trajectory. The AUQ study also reports that confidence at the final step outperforms average confidence on ALFWorld (0.913 vs.\ 0.783 AUROC), but does not explain why intermediate confidence is less informative.

Our work addresses this gap by systematically analyzing whether and when uncertainty metrics become predictive of failure in long-horizon deep research tasks.

\section{Conclusion}

Across the evaluated deep-research benchmarks, verbal confidence reliably detects failures at task completion (mean AUROC 0.85) but has limited predictive ability at intermediate stages, with no evaluated metric exceeding a mean AUROC of 0.60 at 50\% progress; neither additional features nor feature combinations close this gap. Path switching, where agents abandon their current search direction in-trajectory, weakens the relationship between intermediate uncertainty and final outcomes, and helps explain why discriminative signal becomes strong only near completion. Our intervention experiments therefore support a practical recommendation for agent harnesses in deep-research settings: use final-step confidence to decide whether to restart, as confidence-guided post-completion retries outperform in-trajectory intervention.

\newpage
\section*{Limitations}

Our evaluation primarily focuses on long-horizon deep research tasks, with TerminalBench~2.0 providing an additional comparison with coding tasks. The results suggest that uncertainty dynamics vary across task types, potentially depending on whether agent actions modify the environment state and how frequently agents switch paths. Examining these differences across a broader range of tasks, including mathematical reasoning, coding, multi-turn dialogue, shopping, and desktop control, remains an important direction for future work. In addition, our experiments focus on mainstream uncertainty quantification metrics, including verbal confidence, token probability metrics, and consistency-based metrics. Richer trajectory-aware methods and broader harness designs, such as
self-critique and multi-agent monitoring, may enable more reliable early
failure detection and warrant further study.

\section*{Acknowledgments}

This work was supported in part by the National Natural Science Foundation of China (No. 62602580), the Ningbo Youth Science and Technology Innovation Leading Talent Program (No. 2025QL059), the Ant Group Research Fund, and the ``Pioneer and Leading Goose'' R\&D Program of Zhejiang (No. 2025C02037).

\bibliography{references}

\appendix
\input{appendix}

\end{document}

%% file: tab_cross_family.tex
\begin{table*}[t]
\centering
\begin{adjustbox}{max width=\textwidth}
\begin{tabular}{ll rrr rrr rrr rrr rrr}
\toprule
 &  & \multicolumn{3}{c}{\textbf{Qwen3.5-27B}} & \multicolumn{3}{c}{\textbf{Qwen3.5-122B-A10B}} & \multicolumn{3}{c}{\textbf{GLM-4.7}} & \multicolumn{3}{c}{\textbf{DeepSeek-V3.2}} & \multicolumn{3}{c}{\textbf{Kimi K2.6}} \\
\cmidrule(lr){3-5} \cmidrule(lr){6-8} \cmidrule(lr){9-11} \cmidrule(lr){12-14} \cmidrule(lr){15-17}
\textbf{Metric} & \textbf{Best agg.} & \textbf{BC} & \textbf{BC-zh} & \textbf{HLE} & \textbf{BC} & \textbf{BC-zh} & \textbf{HLE} & \textbf{BC} & \textbf{BC-zh} & \textbf{HLE} & \textbf{BC} & \textbf{BC-zh} & \textbf{HLE} & \textbf{BC} & \textbf{BC-zh} & \textbf{HLE} \\
\midrule
\multicolumn{17}{l}{\textit{Single-trajectory}} \\
\midrule
Verbal Confidence~\citep{tian-etal-2023-just} & last step & \textbf{.921} & \textbf{.867} & \textbf{.801} & \textbf{.904} & \textbf{.864} & \textbf{.763} & \textbf{.915} & \textbf{.854} & \textbf{.744} & \textbf{.906} & \textbf{.860} & \textbf{.862} & \textbf{.794} & \textbf{.816} & \textbf{.790} \\
PPL~\citep{jelinek1977perplexity} & last step & \underline{.842} & \underline{.692} & .580 & \underline{.853} & \underline{.752} & .601 & \underline{.816} & \underline{.790} & .692 & .745 & .492 & .481 & .575 & .534 & .547 \\
Entropy~\citep{shannon1948mathematical} & max & .702 & .689 & .667 & .736 & .705 & \underline{.695} & .692 & .748 & \underline{.695} & .750 & .575 & .469 & .607 & .548 & .669 \\
Max NLL~\citep{manakul2023} & max & .692 & .672 & .681 & .700 & .710 & .682 & .720 & .763 & .673 & \underline{.757} & \underline{.631} & .565 & .665 & .622 & .705 \\
Max token entropy~\citep{manakul2023} & max & .717 & .668 & \underline{.706} & .714 & .708 & .679 & .741 & .733 & .690 & .701 & .613 & \underline{.618} & \underline{.761} & \underline{.742} & \underline{.736} \\
\bottomrule
\end{tabular}
\end{adjustbox}
\caption{Best AUROC for each single-trajectory uncertainty metric across 15 combinations of models and benchmarks. The aggregation operator with the most wins across settings is shown for each metric.\textbf{Bold}: best in column; \underline{underlined}: second best. Full results in Appendix~\ref{sec:cross_family_full}.}
\label{tab:cross_family}
\end{table*}

%% file: tab_consistency.tex
\begin{table}[t]
\centering
\resizebox{\columnwidth}{!}{%
\begin{tabular}{l rrr rrr}
\toprule
 & \multicolumn{3}{c}{\textbf{Qwen3.5-122B-A10B}} & \multicolumn{3}{c}{\textbf{GLM-4.7}} \\
\cmidrule(lr){2-4} \cmidrule(lr){5-7}
\textbf{Metric} & \textbf{BC} & \textbf{BC-zh} & \textbf{HLE} & \textbf{BC} & \textbf{BC-zh} & \textbf{HLE} \\
\midrule
\multicolumn{7}{l}{\textit{Single-trajectory}} \\
\midrule
Verbal Confidence~\citep{tian-etal-2023-just} & \textbf{.904} & \textbf{.864} & .763 & \textbf{.915} & \textbf{.854} & .744 \\
\midrule
\multicolumn{7}{l}{\textit{Consistency-based (8 rollouts per task)}} \\
\midrule
Disc.\ sem.\ ent.~\citep{farquhar2024detecting} & .838 & .800 & \textbf{.783} & \underline{.844} & .823 & \textbf{.822} \\
Deg.\ unc.~\citep{lin2024graph} & .849 & \underline{.839} & .748 & .827 & \underline{.839} & .799 \\
Diversity~\citep{cole-etal-2023-selectively} & .840 & .789 & \underline{.778} & .837 & .820 & \underline{.811} \\
Variation ratio~\citep{huang2023look} & \underline{.862} & .814 & .714 & .823 & .817 & .743 \\
Lexical sim.~\citep{fomicheva2020unsupervised} & .861 & .814 & .714 & .823 & .817 & .742 \\
FSD~\citep{lyu2024calibrating} & .804 & .776 & .777 & .801 & .789 & .779 \\
LN pred.\ ent.~\citep{malininuncertainty} & .827 & .716 & .624 & .782 & .772 & .669 \\
Pred.\ entropy~\citep{kadavath2022} & .832 & .667 & .613 & .745 & .745 & .666 \\
SentSAR~\citep{duan2024shifting} & .720 & .685 & .546 & .757 & .733 & .649 \\
Sem.\ entropy~\citep{kuhn2023semantic} & .746 & .709 & .690 & .750 & .576 & .538 \\
\bottomrule
\end{tabular}}
\caption{AUROC at trajectory completion for verbal confidence and consistency-based methods. \textbf{Bold}: best; \underline{underlined}: second best per column.}
\label{tab:consistency}
\end{table}

%% file: appendix.tex
\clearpage
\section*{Appendix}
\addcontentsline{toc}{section}{Appendix}

\section{Sensitivity to Confidence Elicitation}
\label{sec:appendix_prompts}

We evaluate three confidence elicitation methods that differ in where and how
confidence is obtained. Despite differences in elicitation location and
confidence scope, the results show the same qualitative trend across all three
methods.

\paragraph{In-tool-call.}
In our main experiments, the model reports confidence as the final field of
each tool call, after generating its reasoning and the other tool-call
arguments.

\begin{quote}
\small\itshape
An integer between 0 and 100 representing your confidence that this step will yield high-value, relevant information. Think: ``How likely is it that executing this action will retrieve information relevant for the current reasoning step?'' Lower the score (e.g., below 50) if: 1)~The parameters/queries are too broad, vague, or highly speculative; 2)~The expected results are likely repetitive or redundant with existing context; 3)~The action is based on an unverified or weak intermediate hypothesis; 4)~Previous tool calls with similar parameters failed to yield valuable information.
\end{quote}

\paragraph{In-content.}
The model reports its confidence between
\texttt{\textless confidence\textgreater} and
\texttt{\textless/confidence\textgreater} tags after generating its reasoning
and before issuing each tool call.

\begin{quote}
\small\itshape
IMPORTANT PROTOCOL: When you need to call a tool, your output must follow this exact order:
\texttt{<think>...</think>}\allowbreak
\texttt{<confidence>...</confidence>}\allowbreak
\texttt{<tool\_call>...</tool\_call>}.
The \texttt{<confidence>} block is mandatory after every \texttt{</think>}.
The token \texttt{<tool\_call>} must never appear unless it is immediately preceded by a valid
\texttt{<confidence>...</confidence>} block.
Confidence level must be a number between 0 and 100. It represents the probability that continuing along the current solution path will eventually lead to the correct final answer.
\end{quote}

\paragraph{Two-stage.}
Following \citet{tian-etal-2023-just}, confidence is obtained through an additional model call after each step.

\begin{quote}
\small\itshape
You are reviewing a problem-solving trajectory prefix. The trajectory stops immediately after one assistant message. Based only on this prefix, estimate the probability that continuing along this same problem-solving path from this point will eventually produce the correct final answer. Give ONLY the probability, no other words or explanation. For example: Probability: \texttt{<the probability between 0 and 100>}
\end{quote}

\paragraph{Summary.}
Table~\ref{tab:prompt_comparison} summarizes the three methods.

\begin{table}[t]
\centering
\small
\resizebox{\columnwidth}{!}{%
\begin{tabular}{lccc}
\toprule
& \textbf{In-tool-call}
& \textbf{In-content}
& \textbf{Two-stage} \\
\midrule
Confidence location & Final tool field & Content tag & Separate call \\
Elicited during agent execution & Yes & Yes & No \\
\bottomrule
\end{tabular}}
\caption{Comparison of three confidence elicitation methods.}
\label{tab:prompt_comparison}
\end{table}

\section{Full-Trajectory Failure Detection by Metric Family}
\label{sec:cross_family_full}

Table~\ref{tab:cross_family_full} reports full-trajectory AUROC for all single-trajectory metric families and aggregation operators across 15 model and benchmark combinations.

\input{tab_cross_family_full}

\section{Consistency-Based Metric Descriptions}
\label{sec:appendix_consistency_metrics}

Table~\ref{tab:consistency} uses abbreviated metric names; their full names and definitions are provided below.

\paragraph{Discrete semantic entropy (Disc.\ sem.\ ent.).} Responses are grouped into semantic equivalence classes, and uncertainty is the entropy of the empirical class-frequency distribution~\citep{farquhar2024detecting}.

\paragraph{Degree-based uncertainty (Deg.\ unc.).} Responses form nodes in a graph weighted by pairwise semantic similarity, and uncertainty is computed from the normalized degrees of this graph~\citep{lin2024graph}.

\paragraph{Sampling diversity (Diversity).} This metric measures the proportion of distinct answers among sampled responses, with greater diversity indicating greater uncertainty~\citep{cole-etal-2023-selectively}.

\paragraph{Variation ratio.} Variation ratio measures disagreement among sampled responses using pairwise answer similarity~\citep{huang2023look}.

\paragraph{Lexical similarity (Lexical sim.).} Originally denoted D-Lex-Sim, this metric averages pairwise lexical similarity among sampled outputs, with lower similarity indicating greater uncertainty~\citep{fomicheva2020unsupervised}.

\paragraph{First--second distance (FSD).} FSD is the difference between the frequencies of the most common and second-most-common answers, providing a confidence score based on the separation between the two leading answers~\citep{lyu2024calibrating}.

\paragraph{Length-normalized predictive entropy (LN pred.\ ent.).} This metric averages length-normalized negative log-likelihood across sampled responses, reducing the effect of response length on predictive entropy~\citep{malininuncertainty}.

\paragraph{Predictive entropy (Pred.\ entropy).} Predictive entropy estimates the entropy of the answer distribution by averaging the negative log-likelihood of sampled responses~\citep{kadavath2022}.

\paragraph{Sentence-shifted predictive entropy (SentSAR).} SentSAR adjusts the probability of each sampled response using its semantic similarity to other responses before computing predictive entropy~\citep{duan2024shifting}.

\paragraph{Semantic entropy (Sem.\ entropy).} Responses are clustered by semantic equivalence, and uncertainty is computed over the probability mass assigned to the resulting semantic classes~\citep{kuhn2023semantic}.

\section{Predictive Ability by Model and Benchmark}
\label{sec:appendix_norm_auroc_per_dataset}

\begin{figure*}[ht]
\centering
\includegraphics[width=\textwidth]{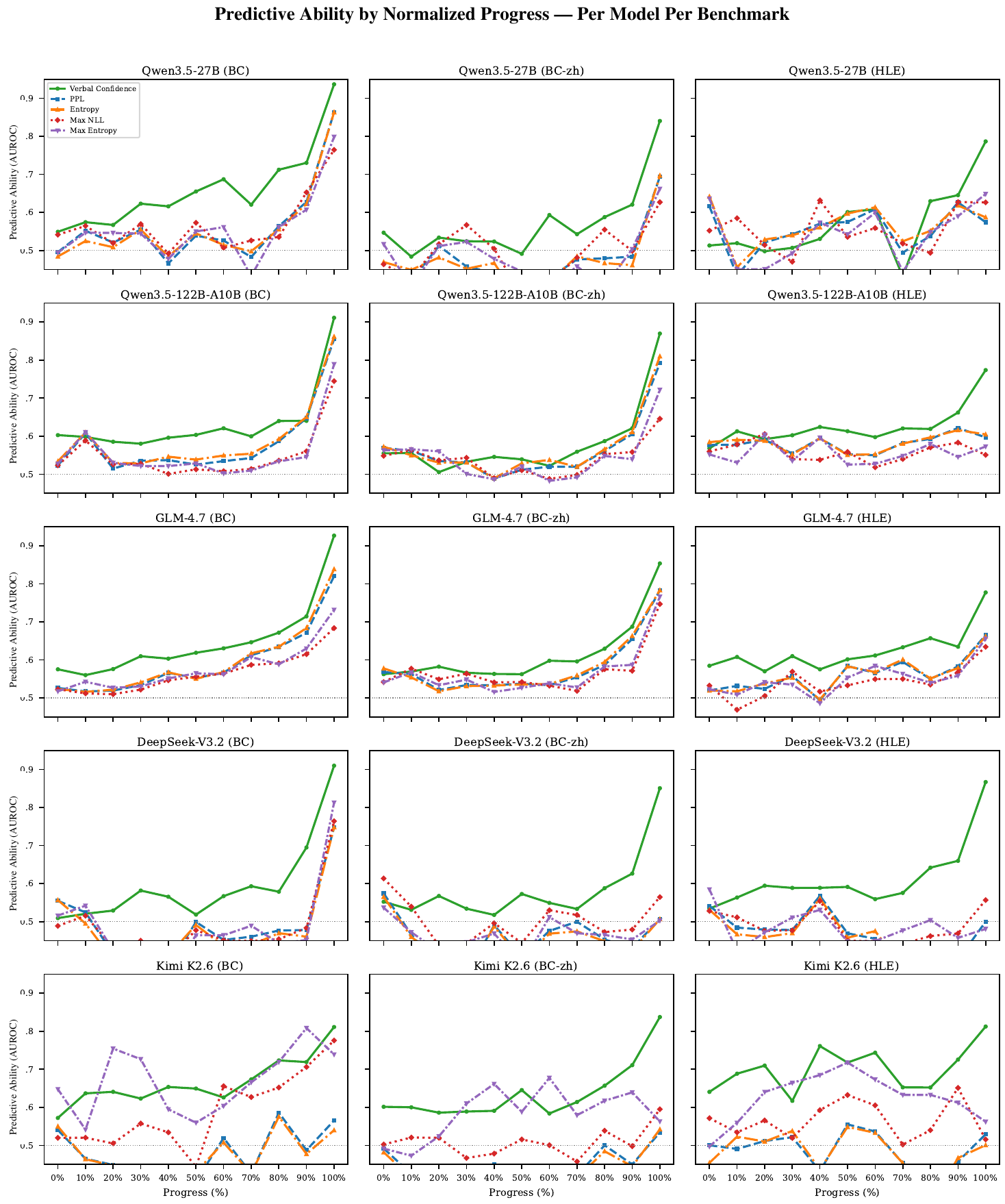}
\caption{Predictive ability of uncertainty metrics for agent failure across normalized trajectory progress for each model and benchmark combination. Each subplot shows five uncertainty metrics: verbal confidence, PPL, entropy, max NLL, and max entropy. Across the 15 settings, predictive ability is limited at intermediate stages and increases near trajectory completion.}
\label{fig:norm_auroc_per_dataset}
\end{figure*}

Figure~\ref{fig:norm_auroc_per_dataset} shows AUROC by normalized trajectory progress for each of the 15 model and benchmark combinations, complementing the averaged results in Figure~\ref{fig:norm_auroc_intro} of the main paper. Across the evaluated settings, uncertainty metrics show limited predictive ability at intermediate stages and become more informative near trajectory completion.

\section{Comparison of Path Switching Between Two Task Types}
\label{sec:appendix_switch_environments}

Table~\ref{tab:switch_environments} compares path switching between
deep-research agents on BrowseComp and coding agents on TerminalBench. Under
the same detection criterion, path switching is less frequent in coding tasks.
This pattern is consistent with the stronger predictive ability of intermediate
uncertainty signals observed in coding tasks.

\begin{table}[t]
\centering
\small
\resizebox{\columnwidth}{!}{%
\begin{tabular}{lrr}
\toprule
\textbf{Setting} & \textbf{Traj. with switch} & \textbf{Avg. switches} \\
\midrule
BrowseComp & 86.8\% & 8.50 \\
TerminalBench & 50.0\% & 2.13 \\
\bottomrule
\end{tabular}%
}
\caption{Comparison of path switching between deep research and coding tasks.}
\label{tab:switch_environments}
\end{table}

\section{Controlling for Task Difficulty}
\label{sec:appendix_switch_difficulty}

The observed relationship between path switching and the low predictive ability
of intermediate uncertainty may be confounded by task difficulty. Harder tasks
may simultaneously cause agents to switch paths more frequently and make early
uncertainty signals less predictive. To control for this potential confounder,
we stratify tasks by difficulty and examine whether the relationship persists
within each group.

We conduct this analysis on BrowseComp trajectories generated by GLM-4.7. We define difficulty using the success rate across eight rollouts for each
task. Tasks with 0--2, 3--5, and 6--8 successful rollouts are categorized as
hard, medium, and easy, respectively. Within each group, we measure the
prevalence of path switching and compare confidence AUROC immediately before
and five steps after the last switch. The results are reported in
Table~\ref{tab:switch_difficulty}.

\begin{table}[t]
\centering
\small
\resizebox{\columnwidth}{!}{%
\begin{tabular}{lrrrrr}
\toprule
\textbf{Difficulty} & \textbf{Acc.} & \textbf{Traj. with switch} &
\textbf{\# Switches} & \textbf{AUROC $t{-}1$} & \textbf{AUROC $t{+}5$} \\
\midrule
Hard   & 4.8\%  & 95.0\% & 10.55 & 0.490 & 0.742 \\
Medium & 50.3\% & 86.4\% & 7.12  & 0.539 & 0.821 \\
Easy   & 90.9\% & 64.1\% & 3.19  & 0.567 & 0.766 \\
\bottomrule
\end{tabular}%
}
\caption{Path switching and confidence AUROC before and after the last switch across task difficulty groups.}
\label{tab:switch_difficulty}
\end{table}

Path switching remains prevalent within all three difficulty groups, although
it is more frequent on harder tasks: 95.0\% of hard, 86.4\% of medium, and
64.1\% of easy trajectories contain at least one switch. More importantly,
confidence becomes more predictive after the last switch within every group.
AUROC increases from 0.490 to 0.742 for hard tasks, from 0.539 to 0.821 for
medium tasks, and from 0.567 to 0.766 for easy tasks. The persistence of this
within-group pattern indicates that task difficulty alone does not explain the
relationship between path switching and predictive ability.

\section{Model Scale and Early Predictive Ability}
\label{sec:appendix_model_scale}

The evaluated models vary in parameter count and overall task performance, raising the question of whether model scale or capability is related to predictive ability at intermediate stages. To examine this relationship, we aggregate verbal confidence AUROC across the three evaluated benchmarks for each model. Table~\ref{tab:model_scale_early_prediction} reports the macro-averaged success rate and AUROC at four stages of trajectory progress.

\begin{table}[t]
\centering
\small
\resizebox{\columnwidth}{!}{%
\begin{tabular}{lrrrrrr}
\toprule
\textbf{Model} & \textbf{Total parameters} & \textbf{Success}
& \textbf{20\%} & \textbf{50\%} & \textbf{70\%} & \textbf{100\%} \\
\midrule
Qwen3.5-27B       & 27B  & 0.411 & 0.533 & 0.583 & 0.532 & 0.855 \\
Qwen3.5-122B-A10B & 122B & 0.394 & 0.561 & 0.585 & 0.593 & 0.851 \\
GLM-4.7           & 355B & 0.384 & 0.576 & 0.594 & 0.625 & 0.852 \\
DeepSeek-V3.2     & 671B & 0.497 & 0.564 & 0.561 & 0.567 & 0.876 \\
Kimi K2.6         & 1T   & 0.624 & 0.645 & 0.671 & 0.647 & 0.820 \\
\bottomrule
\end{tabular}%
}
\caption{Relationship between model size and the early failure-prediction
ability of verbal confidence. AUROC is reported at different trajectory
progress points for each model and averaged across the three benchmarks.
Success denotes the mean task success rate. Model size refers to total rather
than activated parameters.}
\label{tab:model_scale_early_prediction}
\end{table}

The models at the two ends of the parameter count range show a clear difference in early predictive ability. Qwen3.5-27B, the smallest evaluated model, has the weakest early predictive ability, with AUROC values of 0.533 at 20\% progress and 0.583 at 50\%. Kimi K2.6, which has the largest total parameter count and the highest average success rate among the evaluated models, achieves the highest AUROC at the same stages, reaching 0.645 at 20\% progress and 0.671 at 50\%. However, the three intermediate models do not follow a monotonic ordering by either total parameter count or success rate. For example, DeepSeek-V3.2 has a higher success rate than GLM-4.7 but lower AUROC at both 50\% and 70\% progress.

These results show that early predictive ability varies across models, but the present comparison does not establish a systematic relationship with either model scale or task performance. The models also differ in architecture, activated parameter count, and training procedure. Confidence at completion remains more informative than intermediate confidence for every model. Even for Kimi K2.6, which achieves the highest AUROC at 50\% progress, AUROC increases from 0.671 at 50\% to 0.820 at completion.

\section{Early Failure Detection with an External Monitor}
\label{sec:appendix_external_monitor}

The limited early predictive power of standard uncertainty signals suggests
that reliable early failure detection may require richer trajectory-aware
methods or broader harness designs, such as self-critique or multi-agent
monitoring. As an illustrative experiment, we use
GLM-5.1~\citep{glm5team2026glm5vibecodingagentic,zai2026glm51} as an external
monitor for GLM-4.7 trajectories on BrowseComp. At each progress point, the
monitor observes only the trajectory prefix, without access to future steps,
the final answer, or the ground truth, and assesses failure risk from
behavioral patterns such as repeated unsuccessful searches.
Table~\ref{tab:external_monitor} compares its AUROC with that of verbal
confidence throughout trajectory progress.

\begin{table}[t]
\centering
\small
\resizebox{\columnwidth}{!}{%
\begin{tabular}{crr}
\toprule
\textbf{Progress} & \textbf{Verbal confidence} & \textbf{Repeated failed search} \\
\midrule
10\%  & 0.555 & 0.622 \\
30\%  & 0.584 & 0.663 \\
50\%  & 0.560 & 0.680 \\
70\%  & 0.584 & 0.727 \\
90\%  & 0.701 & 0.767 \\
100\% & 0.933 & 0.806 \\
\bottomrule
\end{tabular}
}
\caption{AUROC of verbal confidence and repeated failed searches identified by an external monitor on GLM-4.7 BrowseComp trajectories.}
\label{tab:external_monitor}
\end{table}

The external monitor signal yields higher AUROC than verbal confidence at every evaluated point within the first 90\% of trajectory progress. At 50\% progress, its AUROC is 0.680, compared with 0.560 for verbal confidence. At completion, verbal confidence becomes more informative, reaching 0.933 AUROC compared with 0.806 for the monitor. These results provide initial evidence that trajectory diagnostics can offer useful early failure signals beyond the UQ metrics evaluated in our main experiments. They motivate future work on richer trajectory monitors, critic agents, and architectures with planner and executor components.

\section{Verbal Confidence Trajectories by Model and Benchmark}
\label{sec:appendix_confidence_trajectories}

\begin{figure*}[ht]
\centering
\includegraphics[width=\textwidth]{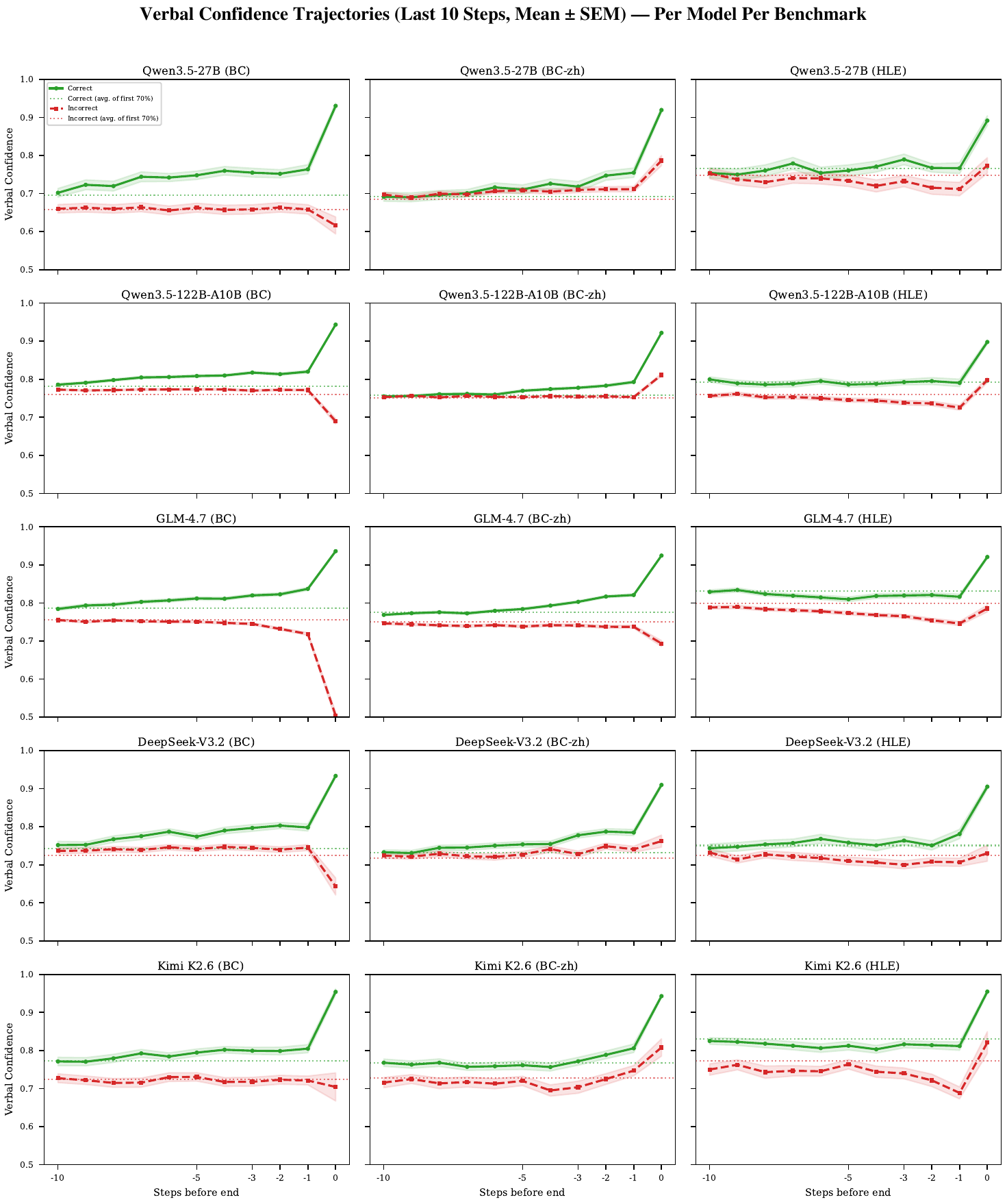}
\caption{Verbal confidence over the last 10 steps for each model and benchmark combination (shaded area: $\pm$1 SEM). Dashed lines indicate the average confidence over the first 70\% of the trajectory for correct (green) and incorrect (red) trajectories. The confidence surge near trajectory completion is observed across all settings, although its magnitude varies.}
\label{fig:confidence_trajectories_full}
\end{figure*}

Figure~\ref{fig:confidence_trajectories_full} shows verbal confidence over the last 10 steps for each model and benchmark combination, complementing the averaged results in Figure~\ref{fig:path_switch_combined}(b) of the main paper.

\section{Failure Detection Precision--Recall Curves}
\label{sec:appendix_pr_curves}

\begin{figure*}[ht]
\centering
\includegraphics[width=\textwidth]{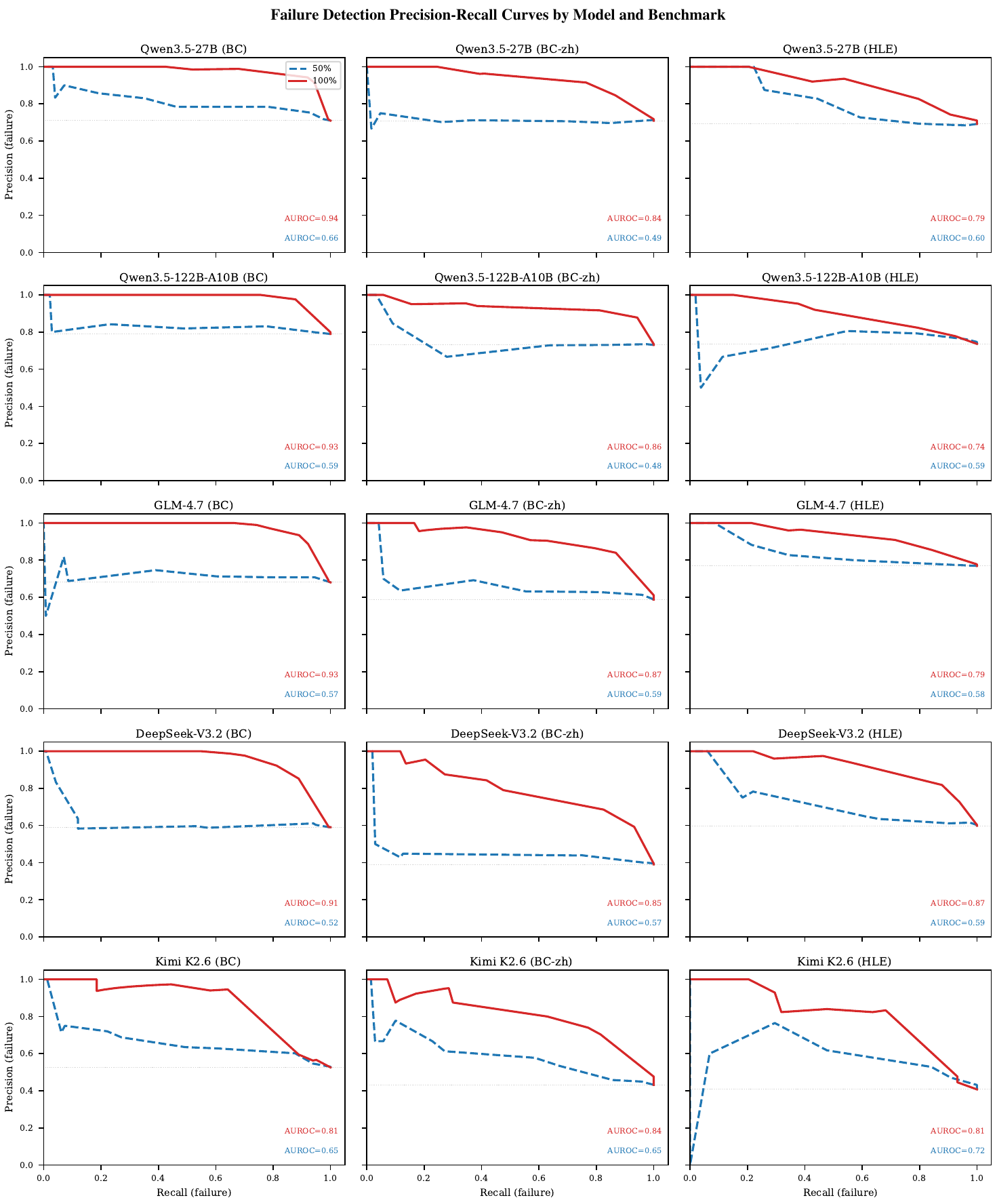}
\caption{Precision--recall curves for failure detection using verbal confidence at 50\% (blue dashed) and 100\% (red solid) trajectory progress. Each subplot corresponds to one model and benchmark combination. At 50\% progress, precision remains close to the failure rate (gray dotted line) across much of the recall range; at 100\% progress, the precision--recall tradeoff improves.}
\label{fig:pr_per_dataset}
\end{figure*}

Figure~\ref{fig:pr_per_dataset} shows precision--recall curves for failure detection using verbal confidence at 50\% and 100\% trajectory progress for each model and benchmark combination. At 50\% progress, precision remains close to the failure rate across much of the recall range, indicating limited discriminative ability at intermediate stages. At 100\% progress, the precision--recall tradeoff improves, showing that confidence becomes more informative near trajectory completion.

%% file: tab_cross_family_full.tex
\begin{table*}[ht]
\centering
\begin{adjustbox}{max width=\textwidth}
\begin{tabular}{ll rrr rrr rrr rrr rrr}
\toprule
 &  & \multicolumn{3}{c}{\textbf{Qwen3.5-27B}} & \multicolumn{3}{c}{\textbf{Qwen3.5-122B-A10B}} & \multicolumn{3}{c}{\textbf{GLM-4.7}} & \multicolumn{3}{c}{\textbf{DeepSeek-V3.2}} & \multicolumn{3}{c}{\textbf{Kimi K2.6}} \\
\cmidrule(lr){3-5} \cmidrule(lr){6-8} \cmidrule(lr){9-11} \cmidrule(lr){12-14} \cmidrule(lr){15-17}
\textbf{Metric} & \textbf{Aggregation} & \textbf{BC} & \textbf{BC-zh} & \textbf{HLE} & \textbf{BC} & \textbf{BC-zh} & \textbf{HLE} & \textbf{BC} & \textbf{BC-zh} & \textbf{HLE} & \textbf{BC} & \textbf{BC-zh} & \textbf{HLE} & \textbf{BC} & \textbf{BC-zh} & \textbf{HLE} \\
\midrule
\multicolumn{17}{l}{\textit{Single-trajectory}} \\
\midrule
\multicolumn{17}{l}{\textit{Verbal Confidence}~\citep{tian-etal-2023-just}} \\
 & last step & \textbf{.921} & \textbf{.867} & \textbf{.801} & \textbf{.904} & \textbf{.864} & \textbf{.763} & \textbf{.915} & \textbf{.854} & \underline{.744} & \textbf{.906} & \textbf{.860} & \textbf{.862} & \textbf{.794} & \textbf{.816} & .790 \\
 & last 3 steps & .878 & .821 & .740 & .845 & .783 & \underline{.740} & \underline{.897} & .825 & \textbf{.749} & \underline{.863} & .781 & \underline{.806} & .790 & .722 & \textbf{.813} \\
 & last 5 steps & .859 & .753 & .722 & .813 & .719 & .714 & .879 & .804 & .732 & .844 & .748 & .794 & .783 & .717 & \underline{.807} \\
 & max & \underline{.917} & \underline{.865} & \underline{.800} & \underline{.865} & \underline{.848} & .731 & .863 & \underline{.834} & .694 & .836 & \underline{.830} & .800 & \underline{.791} & \underline{.794} & .787 \\
 & min & .737 & .650 & .634 & .757 & .652 & .662 & .804 & .724 & .626 & .730 & .618 & .680 & .690 & .672 & .738 \\
 & mean & .753 & .736 & .654 & .739 & .712 & .691 & .764 & .746 & .667 & .682 & .682 & .713 & .703 & .686 & .761 \\
\midrule
\multicolumn{17}{l}{\textit{PPL}~\citep{jelinek1977perplexity}} \\
 & last step & \textbf{.842} & \textbf{.692} & .580 & \textbf{.853} & \textbf{.752} & .601 & \underline{.816} & \textbf{.790} & .692 & \textbf{.745} & .492 & \underline{.481} & \underline{.575} & \underline{.534} & .547 \\
 & last 3 steps & \underline{.794} & .589 & .647 & \underline{.821} & .693 & .626 & \textbf{.830} & \underline{.788} & \underline{.697} & .679 & \underline{.517} & .443 & .532 & .505 & \underline{.573} \\
 & last 5 steps & .738 & .488 & \underline{.673} & .784 & .636 & .610 & .813 & .765 & .689 & .635 & .510 & .448 & .529 & .462 & .527 \\
 & max & .698 & \underline{.669} & .667 & .722 & \underline{.716} & \textbf{.690} & .699 & .755 & \underline{.697} & \underline{.742} & \textbf{.557} & \textbf{.514} & \textbf{.636} & \textbf{.556} & \textbf{.664} \\
 & min & .367 & .401 & .504 & .493 & .516 & .513 & .497 & .522 & .503 & .307 & .290 & .319 & .314 & .421 & .430 \\
 & mean & .671 & .575 & \textbf{.674} & .731 & .693 & \underline{.678} & .731 & .724 & \textbf{.708} & .392 & .273 & .347 & .446 & .486 & .505 \\
\midrule
\multicolumn{17}{l}{\textit{Entropy}~\citep{shannon1948mathematical}} \\
 & last step & \textbf{.841} & \textbf{.697} & .590 & \textbf{.860} & \textbf{.765} & .606 & \underline{.833} & \textbf{.791} & .687 & \underline{.746} & .486 & \underline{.424} & \underline{.553} & \underline{.547} & .517 \\
 & last 3 steps & \underline{.786} & .606 & .658 & \underline{.825} & \underline{.706} & .627 & \textbf{.837} & \underline{.784} & .694 & .659 & \underline{.531} & .415 & .494 & .500 & \underline{.532} \\
 & last 5 steps & .730 & .500 & \underline{.690} & .793 & .653 & .612 & .818 & .756 & .686 & .600 & .512 & .415 & .481 & .458 & .486 \\
 & max & .702 & \underline{.689} & .667 & .736 & .705 & \textbf{.695} & .692 & .748 & \underline{.695} & \textbf{.750} & \textbf{.575} & \textbf{.469} & \textbf{.607} & \textbf{.548} & \textbf{.669} \\
 & min & .373 & .397 & .530 & .517 & .527 & .516 & .509 & .518 & .505 & .355 & .294 & .294 & .313 & .430 & .403 \\
 & mean & .677 & .583 & \textbf{.691} & .743 & .699 & \underline{.675} & .736 & .721 & \textbf{.705} & .376 & .268 & .331 & .435 & .488 & .496 \\
\midrule
\multicolumn{17}{l}{\textit{Max NLL}~\citep{manakul2023}} \\
 & last step & \textbf{.741} & \underline{.603} & .627 & \textbf{.750} & \underline{.654} & .599 & .694 & \underline{.765} & .680 & \textbf{.761} & .547 & \textbf{.565} & \underline{.744} & \underline{.604} & .533 \\
 & last 3 steps & \underline{.702} & .576 & .692 & .690 & .619 & .635 & \underline{.743} & \textbf{.767} & \textbf{.697} & .587 & \underline{.568} & .458 & .742 & .590 & .600 \\
 & last 5 steps & .642 & .535 & \textbf{.750} & .677 & .589 & .627 & \textbf{.754} & .748 & \underline{.684} & .548 & .551 & \underline{.482} & \textbf{.766} & .583 & \underline{.665} \\
 & max & .692 & \textbf{.672} & .681 & \underline{.700} & \textbf{.710} & \textbf{.682} & .720 & .763 & .673 & \underline{.757} & \textbf{.631} & \textbf{.565} & .665 & \textbf{.622} & \textbf{.705} \\
 & min & .350 & .346 & .549 & .413 & .452 & .507 & .418 & .450 & .493 & .269 & .308 & .332 & .454 & .468 & .485 \\
 & mean & .618 & .578 & \underline{.704} & .662 & .652 & \underline{.654} & .683 & .683 & .681 & .349 & .379 & .353 & .688 & .552 & .627 \\
\midrule
\multicolumn{17}{l}{\textit{Max token entropy}~\citep{manakul2023}} \\
 & last step & \textbf{.780} & \underline{.638} & .652 & \textbf{.796} & \underline{.701} & .617 & .736 & \textbf{.778} & .688 & \textbf{.812} & .493 & \underline{.491} & .700 & .573 & .546 \\
 & last 3 steps & \underline{.748} & .573 & .695 & \underline{.746} & .645 & .644 & \underline{.767} & \underline{.741} & \textbf{.699} & .654 & \underline{.592} & .415 & \textbf{.818} & .666 & .679 \\
 & last 5 steps & .700 & .500 & \textbf{.718} & .707 & .620 & .631 & \textbf{.775} & .709 & .677 & .583 & .544 & .450 & \underline{.812} & \underline{.697} & .693 \\
 & max & .717 & \textbf{.668} & \underline{.706} & .714 & \textbf{.708} & \textbf{.679} & .741 & .733 & \underline{.690} & \underline{.701} & \textbf{.613} & \textbf{.618} & .761 & \textbf{.742} & \underline{.736} \\
 & min & .340 & .357 & .568 & .473 & .483 & .531 & .460 & .467 & .494 & .292 & .259 & .324 & .643 & .479 & .542 \\
 & mean & .655 & .527 & .704 & .691 & .667 & \underline{.659} & .713 & .664 & .679 & .399 & .313 & .370 & .754 & .682 & \textbf{.753} \\
\bottomrule
\end{tabular}
\end{adjustbox}
\caption{Full-trajectory AUROC for all single-trajectory metric families and aggregation operators across 15 model and benchmark combinations (companion to Table~\ref{tab:cross_family}). \texttt{max}/\texttt{min} denote running extrema. Within each family, \textbf{bold} marks the best and \underline{underlined} marks the second best in each column.}
\label{tab:cross_family_full}
\end{table*}